\documentclass[times,twocolumn,final,authoryear]{elsarticle}

\usepackage{ycviu}
\usepackage{framed,multirow}
\usepackage{float}

\usepackage{amsmath}
\usepackage{mathtools}
\usepackage{amssymb}
\usepackage[ruled,vlined]{algorithm2e}
\usepackage{tabularx}
\usepackage{booktabs}
\usepackage{caption}
\usepackage{geometry}
\usepackage{graphicx}
\usepackage{subcaption}
\usepackage{afterpage}

\usepackage{amssymb}
\usepackage{latexsym}
\usepackage{float}

\usepackage{url}
\usepackage{xcolor}
\definecolor{newcolor}{rgb}{.8,.349,.1}
\usepackage[colorlinks,citecolor=red,urlcolor=blue,bookmarks=false,hypertexnames=true]{hyperref}

\journal{Computer Vision and Image Understanding}

\begin{document}

\ifpreprint
  \setcounter{page}{1}
\else
  \setcounter{page}{1}
\fi

\begin{frontmatter}

\title{SpATr: MoCap 3D Human Action Recognition based on Spiral Auto-encoder and Transformer Network}

\author[1]{Hamza \snm{Bouzid} \corref{cor1}} 
\cortext[cor1]{Corresponding author
  }
\ead{hamza-bouzid@um5r.ac.ma }
\author[1]{Lahoucine \snm{Ballihi} }
\ead{lahoucine.ballihi@fsr.um5.ac.ma}

\address[1]{LRIT-CNRST URAC 29, Mohammed V University, Faculty Of Sciences, in Rabat, Morocco.}

\received{}
\finalform{}
\accepted{}
\availableonline{}
\communicated{}

\begin{abstract}
Recent technological advancements have significantly expanded the potential of human action recognition through harnessing the power of 3D data. This data provides a richer understanding of actions, including depth information that enables more accurate analysis of spatial and temporal characteristics. In this context, We study the challenge of 3D human action recognition. 
Unlike prior methods, that rely on sampling 2D depth images, skeleton points, or point clouds, often leading to substantial memory requirements and the ability to handle only short sequences, we introduce a novel approach for 3D human action recognition, denoted as SpATr (Spiral Auto-encoder and Transformer Network), specifically designed for fixed-topology mesh sequences. 
The SpATr model disentangles space and time in the mesh sequences. A lightweight auto-encoder, based on spiral convolutions, is employed to extract spatial geometrical features from each 3D mesh. These convolutions are lightweight and specifically designed for fix-topology mesh data.  Subsequently, a temporal transformer, based on self-attention, captures the temporal context within the feature sequence. The self-attention mechanism enables long-range dependencies capturing and parallel processing, ensuring scalability for long sequences.
The proposed method is evaluated on three prominent 3D human action datasets: Babel, MoVi, and BMLrub, from the Archive of Motion Capture As Surface Shapes (AMASS). Our results analysis demonstrates the competitive performance of our SpATr model in 3D human action recognition while maintaining efficient memory usage. The code and the training results will soon be made publicly available at 
\url{https://github.com/h-bouzid/spatr}.

\end{abstract}

\begin{keyword}
\MSC 41A05\sep 41A10\sep 65D05\sep 65D17
\KWD Human action recognition\sep auto-encoder\sep transformer\sep spiral convolution\sep AMASS\sep Geometric deep learning.

\end{keyword}

\end{frontmatter}
\begin{figure*}[ht]
\begin{center}
\includegraphics[scale=0.6]{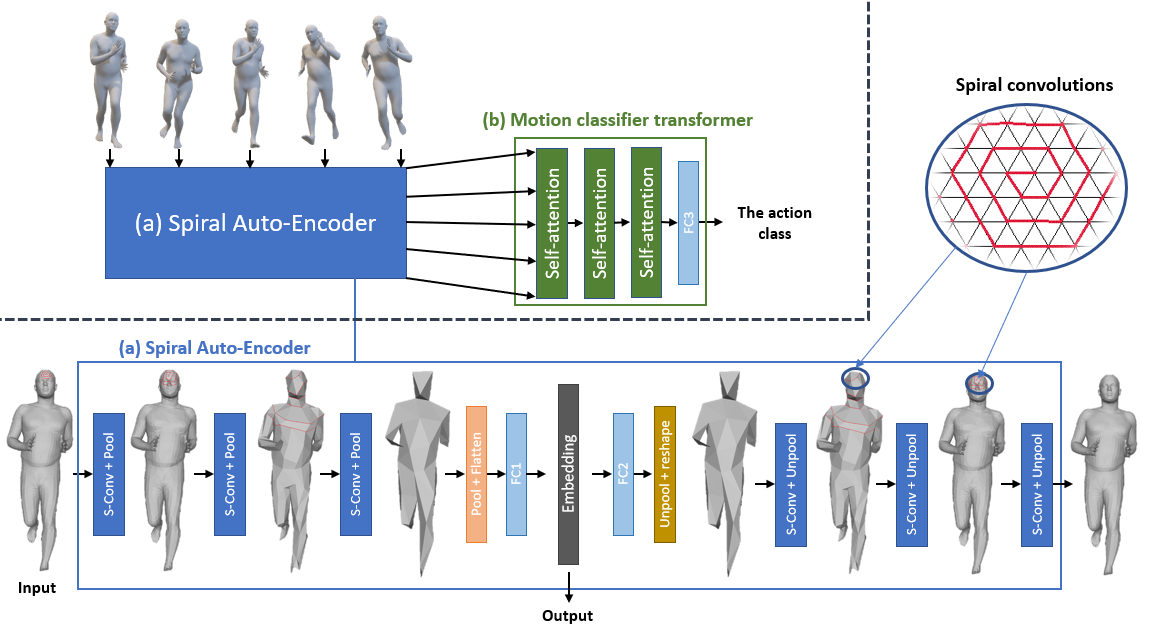}
\caption{Overview of the proposed model for 3D human action recognition. Firstly, each mesh from the action sequence is mapped through the spiral auto-encoder, which extracts its embedding. The spiral auto-encoder is constructed using spiral convolution, and its architecture is illustrated in \textbf{(a)}. Next, the sequence of embeddings is utilized by the temporal transformer, which employs self-attention to capture temporal dependencies between frame embeddings, enabling the prediction of the action class.}
\label{ourApproach}
\end{center}
\end{figure*}
\section{Introduction}

The field of 3D Human Action Recognition (3D HAR) has garnered substantial attention due to rapid advancements in 3D technologies. These technologies hold significant potential across various applications due to the robustness of 3D data against variations in shadowing conditions, lighting, and pose, as well as their capacity to offer precise geometric representations of the human body. However, 3D HAR poses several challenges, including:
\textbf{1)} the complexity of human actions, which can take different motions, speeds, timings, and spatial configurations;
\textbf{2)} the large variability in action classes, making it challenging to train models that generalize well across different actions;
\textbf{3)} the difficulty in finding effective and discriminative representations of 3D data that capture spatio-temporal information, etc.

To address these challenges, various approaches have been proposed. Among them, projection-based methods \citep{chen2016real, li2019action, wang2019generative} have gained significant recognition. These methods involve projecting 3D information onto the 2D domain (depth+texture) and utilizing deep neural networks for classification. Another successful group of methods, known as hand-designed feature descriptor-based methods \citep{hussein2013human,vemulapalli2014human}, utilize handcrafted 3D/4D features extracted from mesh data. Skeleton-based methods \citep{devanne20143,shi2019two,liu2020disentangling,zhang2020semantics,chen2021channel,kacem2018novel} represent human actions using skeletal joint positions and rotations extracted from depth or motion sensors. Point cloud-based methods \citep{li2021sequentialpointnet, fan2022pstnet}, on the other hand, utilize 3D point cloud data to model the human body as a collection of points, thereby capturing spatial and motion information for action recognition. However, the mentioned methods exhibit certain limitations. Projection and skeleton-based methods suffer from the loss of geometrical information. Hand-designed feature descriptor-based methods are expensive, require intense feature selection, and struggle with scalability. Lastly, point cloud-based methods face challenges in regularizing the irregular structure of the human body, spatio-temporal information loss, difficulty in recognizing small and close movements, and, most significantly, substantial complexity and cost associated with these models.\vspace{0.2cm}

Recognizing the potential of motion capture (MoCap) data recognition in various applications, such as sports analysis, animation, gaming, and human-computer interaction, we emphasize the importance of developing dedicated models for MoCap data. Unlike depth data, MoCap data lacks many of the challenges associated with geometric information, spatio-temporal complexities, and irregular body structures. Consequently, it does not necessitate resource-heavy models but rather demands a tailored approach to its unique characteristics. To the best of our knowledge, STMT \citep{zhu2023stmt} is the only state-of-the-art method that achieves 3D HAR by directly using 3D mesh,  without converting it to skeletons, 2D projections, or point clouds. STMT uses a transformer that employs two types of attention mechanisms to capture both spatial and temporal context. However, we argue that the use of attention for both spatial and temporal context extraction can be computationally expensive and might require a large amount of training data, which is not always available. In this context, we propose a 3D HAR lightweight model specifically designed for MoCap fixed-topology data. This model eliminates unnecessary complexity and minimizes resource consumption while ensuring efficient and accurate action recognition. \vspace{0.2cm}

Our proposed model, namely SpATr (Spiral Auto-encoder + Transformer Network), is a fully learned model that directly utilizes mesh data to model the dynamic changes for 3D human action recognition. We decouple spatio-temporal context into spatial and temporal dimensions. First, our model extracts spatial information from each frame in the sequence using an auto-encoder composed of spiral convolutions \citep{bouritsas2019neural}. The sequence of extracted embeddings is then used to train a transformer network \citep{vaswani2017attention} to model the temporal context between frames (inter-frame attention) and classify the 4D motion. The spiral auto-encoder offers task-free embeddings that contain geometrical information and that can be used for different classification tasks (action recognition, identity recognition, anomaly detection, pose estimation, etc.), while the transformer offers high parallelism capabilities and scalability to long sequences with small memory increments, easily reaching thousands of frames. Fig.\ref{ourApproach} illustrates the model overview.\\

In summary, our contributions include the following aspects:
\begin{enumerate}
    \item We propose a fully learned model for 3D human action recognition that directly utilizes 3D human-shaped fixed template meshes as input and predicts the performed action in the sequence.
    \item The proposed model is significantly lightweight compared to the state-of-the-art approaches.
    \item We investigate the impact of decoupling the 3D Human Action Recognition (HAR) process into two distinct models: a spatial spiral auto-encoder and a temporal transformer, and examine the trade-off between model accuracy and memory efficiency.
    \item We extensively evaluate our model on the Babel \citep{punnakkal2021babel}, MoVi \citep{ghorbani2021movi} and BMLrub \citep{troje2002bmldecomposing} datasets for human action recognition from the AMASS archive, and We compare it with the recent 3D HAR state-of-the-art approaches.
\end{enumerate}

\section{Related Work: 3D HAR}
Numerous methods have been proposed to address 3D HAR. These methods involve projection-based approaches, that project 3D information (depth, normal, curvature maps, etc.) to the 2D domain. They then combine this information with 2D views of the object and feed it into classification deep neural networks, typically convolutional networks \citep{yang2012recognizing, chen2016real, wang2019generative}. However, such methods suffer from geometrical information loss during the 2D projection process. Other methods focus on computing hand-crafted 3D/4D features directly from the static/dynamic 3D body scans. Then, they utilize dimensional reduction techniques (e.g., PCA, LDA, NCA) and classic Machine Learning algorithms (e.g., HMMs, SVMs, multi-class Random Forest) for the recognition. The most popular examples of this group of methods include those based on local occupancy patterns and Fourier temporal pyramid \citep{wang2012mining}, covariance matrices \citep{hussein2013human}, rotations and translations between body parts/joints \citep{vemulapalli2014human}. Although hand-designed feature descriptors have shown impressive results and offer compact meaningful insights, they often require intense feature selection algorithms to determine the most relevant features for each task. Moreover, their scalability is limited when dealing with different and larger datasets, increased numbers of action classes, and varying data distributions. Additionally, these methods are primarily designed for short scenes. 

Alternatively,  some researchers have tackled these limitations by utilizing deep learning architectures like RNNs \citep{du2015hierarchical, liu2017skeleton, lee2017ensemble}, temporal CNNs \citep{du2015skeleton, li2017skeleton, ke2017new, hoang20193d}, and GCNs \citep{yan2018spatial, tang2018deep, shi2019two} to model spatial and temporal representations of skeleton sequences. However, these approaches do not fully leverage the available data, as they rely on skeleton representations that lack crucial information, such as topology information, surface motion, and body shape information.\vspace{0.2cm}

To overcome these limitations, another category of methods proposes using geometric deep learning, which studies and mimics deep learning techniques on point clouds, graphs, and manifold data (meshes) in non-euclidean domains. For instance, 3DV-PointNet++ \citep{wang20203dv} employ temporal rank pooling \citep{fernando2016rank} on a set of depth frames to generate a 3D dynamic voxel (3DV) that represents both spatial structure and motion patterns of the sequence. A set of points is then sampled from the 3DV and input into a PointNet++ \citep{qi2017pointnet++} for 3D HAR.
However, forcing a regular voxelized structure on the 3D human body leads to the loss of smoothness (information loss). Additionally, transforming the 3D sequence into a single static point cloud increases computational costs, causes spatio-temporal information loss, and increases the difficulty of recognizing small/close motions. 
In \citep{fan2021point}, the authors address the difficulty of computing accurate point trajectories in point cloud videos, especially colorless point clouds. The proposed model avoids point tracking by employing a point 4D convolution that extracts local patches from each point cloud frame, and a transformer that uses self-attention on the patches to capture position, appearance, and motion information. 
The same authors also propose PSTNet \citep{fan2022pstnet}, where spatial and temporal information is disentangled to model point cloud sequences. PSTNet applies point-based spatial convolution first to extract spatial encoding, followed by temporal convolution to capture temporal information. In addition, due to the difficulty of performing convolution across frames, the point tube technique is proposed to preserve the spatio-temporal local structure. 
In SequentialPointNet \citep{li2021sequentialpointnet}, \textit{Li et al}. propose dividing the main modeling operations into frame-level units executed in parallel. Each of the frames is flattened into a hyperpoint that contains the spatial information. A Hyperpoint-Mixer module is then used to merge the spatial intra-frame and temporal inter-frame features of all hyperpoints to obtain the spatio-temporal representation for HAR. 
Point could methods fail to exploit the connections between vertices, which can offer valuable information. The primary challenge of these methods is the large memory consumption that imposes downsampling the object into a minimal number of points and the sequence into a limited number of frames.
STMT, proposed in \citep{zhu2023stmt}, directly employs 3D MoCap mesh data for 3D human action recognition (HAR), without sampling to skeleton, point cloud, or 2D projection. STMT  utilizes a hierarchical transformer architecture featuring two types of attention mechanisms. The first, intra-frame off-set attention, focuses on spatial information within individual frames. The second, inter-frame self-attention, targets temporal interactions between frames. This dual attention enables grasping the context between any two vertex patches in the spatio-temporal domain.\vspace{0.2cm}

The remainder of this paper is organized as follows: Section \ref{method} describes the proposed method and motivations for using spiral convolutions and transformers. The experimental settings, as well as the quantitative and qualitative analysis of the model, are presented in \ref{experiments}. Finally, a comprehensive conclusion of our work is provided in Section \ref{conclusion}.\vspace{1cm}

\section{Proposed Approach: }
\label{method}
In this section, we present the overall design of our model. Our SpATr is designed to learn recognizing actions performed by a subject in the form of mesh video. The proposed model takes a sequence of MoCap meshes as input and predicts the corresponding action class. Therefore, we formulate our goal as learning a function  $\text{SpATr}(S_{eq}) \rightarrow \Bar{Y}$, where $S_{eq} = [M_1,M_2,...,M_L]$ is the input sequence of length $L$, containing meshes $M_i$ of the same topology, and $\Bar{Y}$ represents the predicted label. \vspace{0.2cm}

The model consists of two networks, a spiral auto-encoder (SpAE) and a temporal transformer classifier ($T$). The SpAE takes in the meshes $M_i$ (frame) of the sequence $S_{eq}$ individually and uses spiral convolutions, proposed in \citep{bouritsas2019neural}, to extract a task-free spatial embedding that represents the mesh, $\text{SpAE}~(M_i)~\rightarrow~E_i$. The transformer $T$ network utilizes self-attention on the sequence of extracted embeddings to learn modeling the temporal evolution of the sequence and recognizing the action, $T(\text{SpAE}(M_1), \text{SpAE}(M_2), ..., \text{SpAE}(M_L)) = T(E_1, E_2, ..., E_L) \rightarrow \Bar{Y}$. The overview of our approach is illustrated in Fig.\ref{ourApproach}.

\subsection{Spiral Auto-Encoder}
\label{SAE_desc}
The primary objective of the spiral auto-encoder (SpAE) lies in finding a mapping denoted as $\text{SpAE} : M_i\in(R^{N\times3}) \rightarrow E_i\in(R^{C})$ such that $M_i \approx \Bar{M_i}$. Conventional techniques for acquiring shape embeddings often utilize linear functions such as PCA or LDA, which causes difficulties in capturing extreme and non-linear deformations. This necessitates the use of deep learning techniques tailored specifically for 3D data. However, developing convolution-like techniques on graphs and manifolds is challenging due to the absence of a comprehensive global coordinate system capable of effectively representing points and their ordering. Earlier intrinsic mesh techniques, such as Geodesic CNN \citep{masci2015geodesic} and Anisotropic CNN \citep{boscaini2016learning}, achieve convolution on mesh data through local systems of coordinates centered around individual vertices. Nonetheless, they lack global context, demand high computational complexity and parameter count, can present optimization challenges, and often require manual feature engineering and pre-computation of local systems of coordinates.\vspace{0.2cm}

We suggest using spiral convolutions, proposed in \citep{bouritsas2019neural}, as a building block of a fully differentiable encoder that learns hierarchical representations $E_i$ of given meshes $M_i$, and a decoder that learns to reconstruct the meshes $\Bar{M_i}$ from the representations $E_i$.
Spiral convolutions are advantageous in taking into account the connectivity of the mesh graph to design global (across the entire mesh) and local (within patches) ordering of the vertices. This enables local processing while capturing the global context for each shape. In addition, spiral filters benefit from weight sharing, which reduces the number of parameters, resulting in a lightweight and easy-to-optimize network.\vspace{0.2cm}

The encoder takes a colorless mesh $M_i$ of $N$ vertices as input. $M_i$ undergoes four layers of spiral convolution, each followed by downsampling. The mesh is then encoded into a $C$-dimensional latent space using a fully connected layer (FC1) to obtain the embedding $E_i$. This embedding is used as input to the decoder, where it passes through a fully connected layer (FC2) and four spiral convolutional layers, each followed by upsampling. The decoder aims to reverse the effect of the encoder and reconstruct the initial mesh $\Bar{M_i}$. 

The network is trained to minimize the L1 distance between the input and the predicted meshes,

\begin{equation}
\label{eq:AEopt}
    \text{SpAE}^* = arg \ \underset{\text{SpAE}}{min} \ \mathcal{L}_{\text{SpAE}}(M_i, \Bar{M_i})\ =\ arg \ \underset{\text{SpAE}}{min} \  [\ \vert \vert M_i\ -\ \Bar{M_i}\ \vert \vert _{1}\ ].
\end{equation}

Reconstructing a mesh similar to the input enables the extraction of features representing the whole object, including shape, expression, pose, and appearance.

\subsection{Temporal Transformer}

Inspired by the success of transformers \citep{vaswani2017attention} in natural language processing \citep{devlin2018bert,yang2019xlnet,brown2020language} and computer vision tasks \citep{wang2021not,arnab2021vivit,zhao2021point}, We leverage self-attention to learn the temporal context of meshes sequences. The self-attention mechanism offers the advantage of parallel training, as each token in the input sequence can be processed independently, and it proves effective in modeling long-range dependencies.\vspace{0.2cm}

Given $E = [E_1, E_2, ..., E_L]$, the embedding sequence that encodes spatial information of the action mesh sequence, we employ self-attention (SA) to learn the temporal context between frames. More precisely, each embedding is used as a token to the transformer. These tokens are then linearly transformed to generate $(Q, K, V)$ the queries, keys, and values, respectively. 
\begin{equation}
    Q = W_q \cdot E, \ K = W_k \cdot E, \ V = W_v \cdot E ,
\label{eq:qkv}
\end{equation}

where $W_q, W_k \in \mathbb{R}^{C_k\times C}$, and $W_v\in \mathbb{R}^{C_v\times C}$ are the weights of the linear transformation, with $C_k$ and  $C_v$ are the dimensions of key and value, respectively. Next, the dot product of the query with all the keys, followed by the softmax function are applied to obtain the  attention weights, 
\begin{equation}
     attention(Q, K) = softmax(\frac{Q^T \cdot K}{\sqrt{C_K}}).
\label{eq:attention}
\end{equation}
These weights represent affinities between different frames, i.e. the temporal context. Finally, $self\mbox{-}attention$ is computed as the weighted sum of the values, where the weights are the learned $attention(Q, K)$.
\begin{equation}
     F_{out} = self\mbox{-}attention(Q, K, V) = V \cdot attention(Q, K)
\label{eq:self-attention}
\end{equation}

To enhance the transformer learning ability, we employ multi-head self-attention, which performs $h$ self-attentions with independent $W_q$, $W_k$, and $W_v$, instead of a single self-attention. The outputs of the self-attention heads are concatenated to construct the output features of a self-attention layer. This multi-head self-attention layer is employed three times in our model. Subsequently, we utilize a fully connected layer (FC3) for action classification.

The transformer is trained to minimize the multi-class cross-entropy loss function, 

\begin{equation}
    T^* = arg \ \underset{T}{min} \ \mathcal{L}_{T}(\mathbf{y}, \mathbf{\Bar{y}})\ =\ arg \ \underset{T}{min} \  [- \sum_{i=1}^{n_c} y_i \log(\Bar{y_i})],
\label{eq:Tloss}
\end{equation}

where $\mathbf{y}$ represents the true label of the input sequence, $\mathbf{\Bar{y}}$ represents the predicted probability distribution over the classes, and $n_c$ is the number of classes.

\subsection{Training Algorithm}
Algorithm \ref{ourAlgoSpAE} outlines the training procedure for the Spiral Auto-Encoder (SpAE). The training dataset (M), comprising $n1$ individual meshes, is input into the algorithm. SpAE takes input meshes, individually, extracts their embeddings using the SpAE encoder \textbf{(4)}, decodes the embeddings to reconstruct the original meshes \textbf{(5)}, and calculates the reconstruction loss \textbf{(6)}. The model parameters are updated using gradient descent to minimize the reconstruction loss \textbf{(7)}. This process is repeated for the specified number of epochs.\\

Algorithm \ref{ourAlgoT} describes the training of the temporal transformer $T$. The training dataset ($S_{eq}(E), Y$), consisting of $n2$ embedding sequences and their corresponding action labels, is fed to the algorithm. Each embedding of the sequence serves as a single token for the Transformer $T$ \textbf{(4)}. The input embedding sequences are fed to the transformer $T$ to obtain predicted labels $\Bar{Y}$ \textbf{(5)}. We then calculate the cross-entropy loss between the predicted $\Bar{Y_i}$ and the target $Y_i$ labels \textbf{(6)}. The loss is backpropagated, and the parameters of Transformer $T$ are updated using gradient descent \textbf{(7)}. These steps are repeated for the specified number of epochs.

\begin{algorithm}[ht]
\SetAlgoLined
\textbf{Input}: Input meshes $M = [M_1,M_2,...,M_{n1}]$;\newline
\textbf{Output}: Output embeddings $E = [E_1,E_2,...,E_{n1}]$; \newline
\nl $\text{SpEA} \ \longleftarrow \ $ initialized with learnable parameters sampled from a Gaussian distribution;\newline
\nl \For{ number of epochs} {
    \nl \For{the number of iterations in epoch}{
        \nl $E_i \ \longleftarrow \ Encoder_{\text{SpAE}}(M_i)$;\newline
        \nl $\Bar{M_i} \ \longleftarrow \ Decoder_{\text{SpAE}}(E_i) $; \newline 
        \nl $\mathcal{L}_{rec} \ \longleftarrow \vert \vert \ M_i\ -\ \Bar{M_i}\ \vert \vert _{1}$;\newline
        \nl $\text{SpEA} \ \longleftarrow \text{SpEA} - \alpha (\partial \mathcal{L}_{rec} / \partial \text{SpEA});$
 } } 
 \caption{Spiral auto-encoder SpAE training algorithm.}
 \label{ourAlgoSpAE}
\end{algorithm}

\begin{algorithm}[ht]
\SetAlgoLined
\textbf{Input}: Input Embedding Sequences $E(S_{eq}) = [E(S_{eq-1}),E(S_{eq-2}),...,E(S_{eq-n2})] = [[E_1,E_2,...,E_l]_1,[E_1,E_2,...,E_l]_2,...,[E_1,E_2,...,E_l]_{n2}]$;\newline
\textbf{Output}: The sequences actions labels $\Bar{Y} = [\Bar{Y}_1,\Bar{Y}_2,...,\Bar{Y}_n2]$; \newline
\nl $ T \ \longleftarrow \ $ initialized with learnable parameters sampled from a Gaussian distribution;\newline
\nl \For{ number of epochs} {
    \nl \For{the number of iterations in epoch}{
        \nl  $T_{Tokens} \ \longleftarrow \ [E_1,E_2,...,E_l]_i$;\newline
        \nl $\Bar{Y_i} \ \longleftarrow \ T(T_{Tokens}) $; \newline 
        \nl $\mathcal{L}_{CrossEnt} \ \longleftarrow  \ -\sum_{i=1}^{C} Y_i \log(\Bar{Y}_i)$;\newline
        \nl $T \ \longleftarrow T - \alpha (\partial \mathcal{L}_{CrossEnt} / \partial T);$
 } } 
 \caption{Temporal transformer T training algorithm.}
 \label{ourAlgoT}
\end{algorithm}

\section{Experiment}
\label{experiments}
In this section, we present a comprehensive evaluation of our model and compare it with state-of-the-art models. Firstly, we describe the experimental setup used for our training. Next, we compare our results to the recent state-of-the-art models, including P4Transformer \citep{fan2021point}, PSTNet \citep{fan2022pstnet}, SequentialPointNet \citep{li2021sequentialpointnet}, and STMT \citep{zhu2023stmt}. We then conduct an ablation study to demonstrate the significance of different components in our model. Finally, we discuss both the strengths and weaknesses of our model and analyze the reasons behind them.
\subsection{Datasets}

The evaluation of our method is performed on three MoCap datasets, namely Babel, BMLrub, and MoVi, from the AMASS Archive \citep{mahmood2019amass}. AMASS provides high-quality 3D human MoCap data that is aligned using SMPL model \citep{SMPL:2015}.\\

\textbf{Babel} \citep{punnakkal2021babel}: stands out as one of the largest 3D MoCap databases, combining data derived from 15 datasets from the AMASS archive. Babel includes 43.5 hours of recordings that encapsulate over 250 distinct action categories, performed by more than 346 subjects. In our study, we select the most dominant 60-class subset, resulting in 21,653 sequences with single-class labels. These sequences exhibit a diverse range of frames/objects, varying from a few to thousands of frames, with each object composed of 6,980 vertices. \\

\textbf{BMLrub} \citep{troje2002bmldecomposing}: contains sequences of 111 individuals, comprising 50 males and 61 females, performing twenty actions, in a total of $3061$ actions performed over a duration of $522.69$ minutes. Each sequence consists of 200 to 2500 human-shaped objects (frames) of $6980$ vertices. In this work, we remove two actions that are represented by less than 3 samples.\\

\textbf{MoVi} \citep{ghorbani2021movi}:
contains sequences of 86 individuals, comprising $29$ males and $57$ females, performing a collection of $20$ predefined everyday actions and sports movements,  along with one self-chosen movement. The dataset includes a total of $1864$ actions performed over a duration of $174.39$ minutes. Each sequence consists of an average of 150 to 2000 human-shaped objects (frames) of $6980$ vertices. We remove the "random" self-chosen movement for each subject in our analysis.

\subsection{Implementation Details}
Prior to training the model, we perform a subject-independent data split into training and testing sets, with an 80:20 ratio, for both MoVi and BLMrub datasets. As for Babel, we use the exact same split in \citep{zhu2023stmt}, specifically at the level of individual data instances, in which the data is split into training, validation, and testing sets, with a 70:15:15 ratio, for fair comparison.\\

For the spiral auto-encoder SpAE, we follow the configuration described in prior work by \citep{bouritsas2019neural}. The encoder of SpAE consists of four spiral convolutional layers, with filter sizes of $[16, 32, 64, 128]$, followed by downsampling. Subsequently, a fully connected layer transforms the codes into an embedding vector of size $1024$. The decoder is built using five spiral convolutional layers, with filter sizes of $[128, 64, 32, 32, 16]$, followed by upsampling. All activation functions are Exponential Linear Units (ELUs) \citep{clevert2015fast}. In order to optimize Eq \ref{eq:AEopt}, the network is trained for $300$ epochs using gradient descent and Adam optimizer \citep{kingma2014adam}, with a learning rate of $10^{-3}$ and a decay rate of $0.99$. Weight decay $5 \times 10^{-5}$ is applied to improve the generalization performance of the network. The implementation of the network is carried out with PyTorch Framework, based on the implementation of \citep{bouritsas2019neural}. The training process takes approximately $10$ hours on both MoVi and BMLrub datasets, while it takes approximately $26$ hours on Babel dataset, using an Nvidia Titan V GPU (12GB of memory). \vspace{0.2cm}
\begin{figure}[t]
\begin{center}
\includegraphics[scale=0.5]{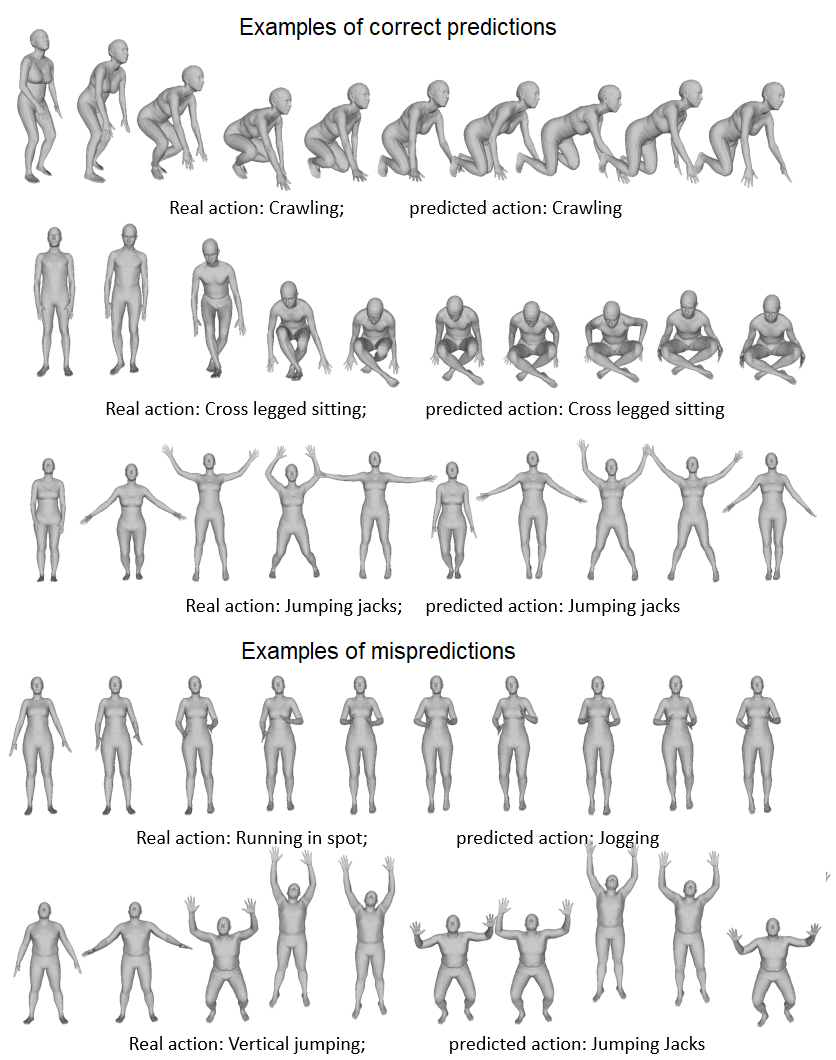}
\caption{Examples of correct and faulty predictions.}
\label{fig:ExamplesOfPrediction}
\end{center}
\end{figure}
The temporal transformer $T$ takes as input a sequence of $L$ embedding vectors of size $1024$, then uses each one of them as a single token. It consists of 3 multi-head self-attention layers with $[2,2,2]$ heads. In order to optimize Eq \ref{eq:Tloss}, we train the transformer for $100$ epochs with gradient descent and Adam optimizer, with a learning rate of $10^{-4}$ and a decay rate of $0.7$. The implementation of the network is conducted with PyTorch Framework. Using an Nvidia GTX 1650 (4GB of memory), the training takes approximately 1 hour on MoVi and BMLrub datasets and about 3 hours on Babel. \\

We established baseline performance by training P4Transformer \citep{fan2021point}, PSTNet \citep{fan2022pstnet}, and SequentialPointNet \citep{li2021sequentialpointnet} using the original methodologies, involving point cloud frame extraction from mesh frames in the dataset and using the official implementations and parameters. For STMT, 2s-AGG-FL, 2s-AGCN-CE, and CTR-GCN, we directly utilized results from the paper by \citep{zhu2023stmt}, as we replicated their evaluation protocol and data partitioning with precision, down to the instance level. This approach ensures a robust basis for meaningful comparisons, aligning our model performance with the benchmarks in a consistent and reliable manner.

\begin{table*}[ht]
\centering
\caption{Action recognition accuracy (\%) results on MoVi BMLrub, AND Babel datasets. }
  \resizebox{0.95\textwidth}{!}{

\begin{tabular}{p{5cm}|c|c|c|cc}
\hline
\hline
 & & \textbf{MoVi} & \textbf{BMLrub} & \multicolumn{2}{c}{\textbf{Babel}} \\
 \textbf{Method} & \textbf{Input}   & Top-1(\%) & Top-1(\%) & Top-1(\%) & Top-5(\%) \\
\hline
\hline
2s-AGCN-FL (CVPR’19) & 3D Skeleton & - & - & 49.62 &  79.12 \\
2s-AGCN-CE  (CVPR’19) & 3D Skeleton & - & - & 63.57 &  86.77 \\
CTR-GCN (ICCV’21) & 3D Skeleton & - & - & 67.30 & 88.50 \\
MS-G3D (CVPR’20) & 3D Skeleton & - & - & 67.43 &  87.99 \\
\hline
P4Transformer (CVPR’21)  & Point cloud & 91.25 & 81.24 & 63.54 & 86.55 \\
PSTNet (ICLR’21) & Point cloud & 88.75 & 80.20 & 61.94 & 84.11 \\
SequentialPointNet (arXiv’21) & Point cloud & \textbf{98.84} & 82.45 & 62.92 & 84.58 \\
\hline
STMT (CVPR’23) & Mesh & - & - & 67.65 &  88.68 \\
\textbf{SpATr (Ours)} & Mesh & 95.42 & \textbf{86.99} & \textbf{70.36} &  \textbf{89.12} \\
\hline

\end{tabular}}
\label{tab:compare_res}
\end{table*}
\begin{table*}[ht]
\centering
\caption{Action recognition accuracy (\%) and memory usage/bach on MoVi dataset.}
  \resizebox{\textwidth}{!}{

\begin{tabular}{p{4cm}|c|c|c|cc}
\hline
\hline
\textbf{Method} & \textbf{Input} & \textbf{nb Frames} & \textbf{nb points} & \textbf{Memory (MB) / batch size }  & \textbf{Accuracy (\%)} \\
\hline
\hline
& & 12 &  & 1184/4 & 89.16 \\
P4Transformer & Point cloud & 24 & 2048 & 2298/4 & 91.25 \\
& & 48 & & 3224/2 & 90.00 \\
\hline
& & 12 & & 2129/2 & 88.33 \\
PSTNet & Point cloud & 24 & 2048 & 3206/2 &  88.75 \\
& & 48 & & 3151/1 & -- \\
\hline
& & 12 &  & 4352/8 & 98.46 \\
SequentialPointNet & Point cloud & 24 & 2048 & 11806/8 & 98.84 \\
& & 48 &  & 40105/8 & 99.22 \\
\hline
& &  12 &  & 480+157/8 & 93.75 \\
& &  24 &  & 480+167/8 & 94.16 \\
 &   & 48 &  & 480+187/8 & 95.42 \\
Our model& Mesh &  96 & 6890 & 480+230/8 & 95.00 \\
& &  192 & & 480+328/8 & 94.58 \\
& &  384 & & 480+554/8 & 94.17 \\
& &  960 & & 480+1480/8 & 94.58 \\
\hline
\end{tabular}}
\label{tab:acc_mem_res}
\end{table*}
\subsection{Experimental Results}

After training our model, we predict the action classes of scenes from the testing subsets. Fig.\ref{fig:ExamplesOfPrediction} presents examples of both successful and failed prediction attempts performed by our proposed model. Notably, we achieve a mean accuracy of 70.36\%, 95.42\%, and 86.99\% for Babel, MoVi, and BMLrub respectively. Fig.\ref{fig:confusion_matrices} displays the performance across action classes of \ref{subfig:movi_res} MoVi, \ref{subfig:bml_res} BMLrub, and \ref{subfig:babel_res} Babel, visualized as confusion matrices (the figure is presented in the Appendix section due to its large size). The figure indicates that our model provides favorable results on MoVi and BMLrub datasets. However, the Babel dataset poses a challenge due to its inherent characteristics, resulting in comparatively lower accuracy. This challenge is attributed to the considerable class imbalance in Babel. The training set encompasses class instances ranging from as few as 6 to a substantial 3000 instances, creating a highly skewed distribution (long-tailed dataset), which makes it difficult for the model to achieve high recognition rates.\vspace{0.2cm}

Next, we compare our approach with the state-of-the-art in terms of max achieved accuracy, using Babel, MoVi, and BMLrub datasets. The baselines encompass mesh, point cloud-based, and skeleton-based methods. The accuracy results of our model and the baselines across all datasets are presented in table \ref{tab:compare_res}. Notably, our method demonstrates competitive performance compared to the baselines. It achieves the highest accuracy on the Babel and BMLrub datasets and secures the second-highest accuracy on the MoVi dataset.\vspace{0.2cm}

In addition, we conduct a follow-up study in which we compare model resource consumption to prove the efficiency and scalability of our model. We compare our proposed model and baselines: P4Transformer, PSTNet, and SequentialPointNet on the MoVi dataset. We assess their performance based on accuracy and memory usage across different frame counts. We note that STMT is not included in this comparison due to the unavailability of its source code. The results of this comparison are illustrated in table \ref{tab:acc_mem_res}. The results reveal that our model shows a higher recognition rate ($94.16\%$) than P4Transformer ($91.25\%$) and PSTNet ($88.75\%$), while it falls behind SequentialPointNet ($98.84\%$). However, it is noteworthy that our model uses significantly less memory compared to the baselines. For instance, when trained on 24 frames our method utilizes a total of 647 MB of GPU memory for a batch size of 8, while P4Transformer uses 2298MB for a batch size of 4, PSTNet uses 3206 MB for a batch size of 2, and SequentialPointNet uses 10265 MB for a batch size of 8. In addition, our model uses the whole 6890 vertices of the objects without any downsampling, whereas the baselines use subsamples of 2048 points. Moreover, Our model demonstrates high scalability by handling large sequences effectively, while the baselines experience difficulties with large memory consumption when increasing the number of frames. For example, SequentialPointNet requires more than 40GB of memory when trained on 48 frames, while ours can reach 960 frames with approximately 2GB.\vspace{0.2cm}
 \begin{table*}[ht]
\centering
\caption{Performance comparison of our model when changing the temporal transformer with MLP, LSTM, and CNN.}
  \resizebox{\textwidth}{!}{

\begin{tabular}{c|cc|cc|cc|cc}
\hline
\textbf{Model}& \multicolumn{2}{c}{with MLP} & \multicolumn{2}{c}{with LSTM} & \multicolumn{2}{c}{with CNN} & \multicolumn{2}{c}{Our model}  \\
\hline
\textbf{nb frames} & \textbf{Acc (\%)} & \textbf{Mem (MB)} & \textbf{Acc (\%)} & \textbf{Mem (MB)}& \textbf{Acc (\%)} & \textbf{Mem (MB)}& \textbf{Acc (\%)} & \textbf{Mem (MB)}  \\\hline
\hline
 24 & 94.58 & 585 & 95.42 & 25 & 88.75 & 12 & 94.14 & 167 \\
 48 & 95.00 & 1192 & 94.58 & 29 & 91.25 & 69 & 95.42 & 187 \\
 96 & 92.92 & 2315 & 93.75 & 40 & 89.58 & 93 & 95.00 & 230 \\
 192 &  \multicolumn{2}{c|}{Out Of Memory ($\ge$ 4GB)}  & 90.00 & 59 & 92.50 & 184 & 93.75 & 328 \\
 960 &  \multicolumn{2}{c|}{Out Of Memory ($\ge$ 4GB)}  & 74.17 & 212 & 87.04 & 488 & 94.58 & 1480 \\
\hline
\end{tabular}}
\label{tab:ablation-res}
\end{table*}

The findings of this analysis suggest that our model performs on par with state-of-the-art methods in terms of recognition rate while exhibiting significantly better performance in terms of memory efficiency and handling large sequences.\vspace{0.2cm}

We note that the batch size varies between setups due to the differences in the maximum capacity of the GPUs used to train the models (our model, P4Transformer, and PSTNet were trained on NVIDIA GeForce GTX 1650 4GB, while SequentialPointNet was trained on NVIDIA Quadro RTX 8000 48GB).

\subsection{Ablation Study}
In this section, our objective is to demonstrate the significance of self-attention in capturing temporal context and to compare it to other techniques. This is achieved by conducting an ablation study on the MoVi dataset. We train different versions of our model, where the temporal transformer is replaced by other networks, in order to observe their impacts in terms of recognition proficiency and memory efficiency. 
Specifically, we train three versions of the proposed model: one with a Multi-Layer Perceptron (MLP) containing 3 hidden layers, another with a Long Short-Term Memory (LSTM) network comprising 3 LSTM layers, and the third with a Convolutional Neural Network (CNN) containing 3 convolutional layers.  All networks are trained multiple times using sequences of different lengths {24,48,96,192,960}. We maintain the same spiral auto-encoder, parameters, losses, and number of epochs for training all models. The accuracy and memory consumption for the trained models is outlined in table \ref{tab:ablation-res}, thereby facilitating a comparative analysis with the complete model.

The results reveal the following findings:
\begin{itemize}
    \item In comparison to the MLP-based model, our proposed model demonstrates competitive accuracy while being more memory efficient. This is attributed to the dense connectivity and larger parameter count of MLP networks. 

    \item Although the LSTM-based model exhibits lower memory consumption and comparable performance to our model for short sequences, it falls short when applied to larger sequences due to difficulties in capturing long-term dependencies. 

    \item While the CNN-based model demonstrates better memory efficiency than our model, its accuracy significantly lags behind. This arises from the limited capability of CNNs in capturing global dependencies and long-range interactions in the data. 

\end{itemize}

Consequently, the transformer-based model emerges as the most favorable choice, showcasing superior performance in terms of accuracy and the ability to capture long-range dependencies, while still maintaining reasonable memory efficiency compared to the other networks considered in this study.

\subsection{Retraining The Model For Other Tasks}
We also showcase the versatility of features extracted from the spiral auto-encoder, emphasizing their applicability across different tasks without requiring retraining. Leveraging these task-free features, we employ various classifiers to predict subject actions, identities, and sexes. We achieve $95.42\%$ accuracy for action classification, $100\%$ for identity recognition, and $100\%$ for sex prediction. These outcomes confirm the ability of the spiral auto-encoder to generate versatile features, leading to optimized resource allocation efficiency and time savings.

\subsection{Parameters Analysis}
\begin{figure}[h]
\begin{center}
\includegraphics[scale=0.15]{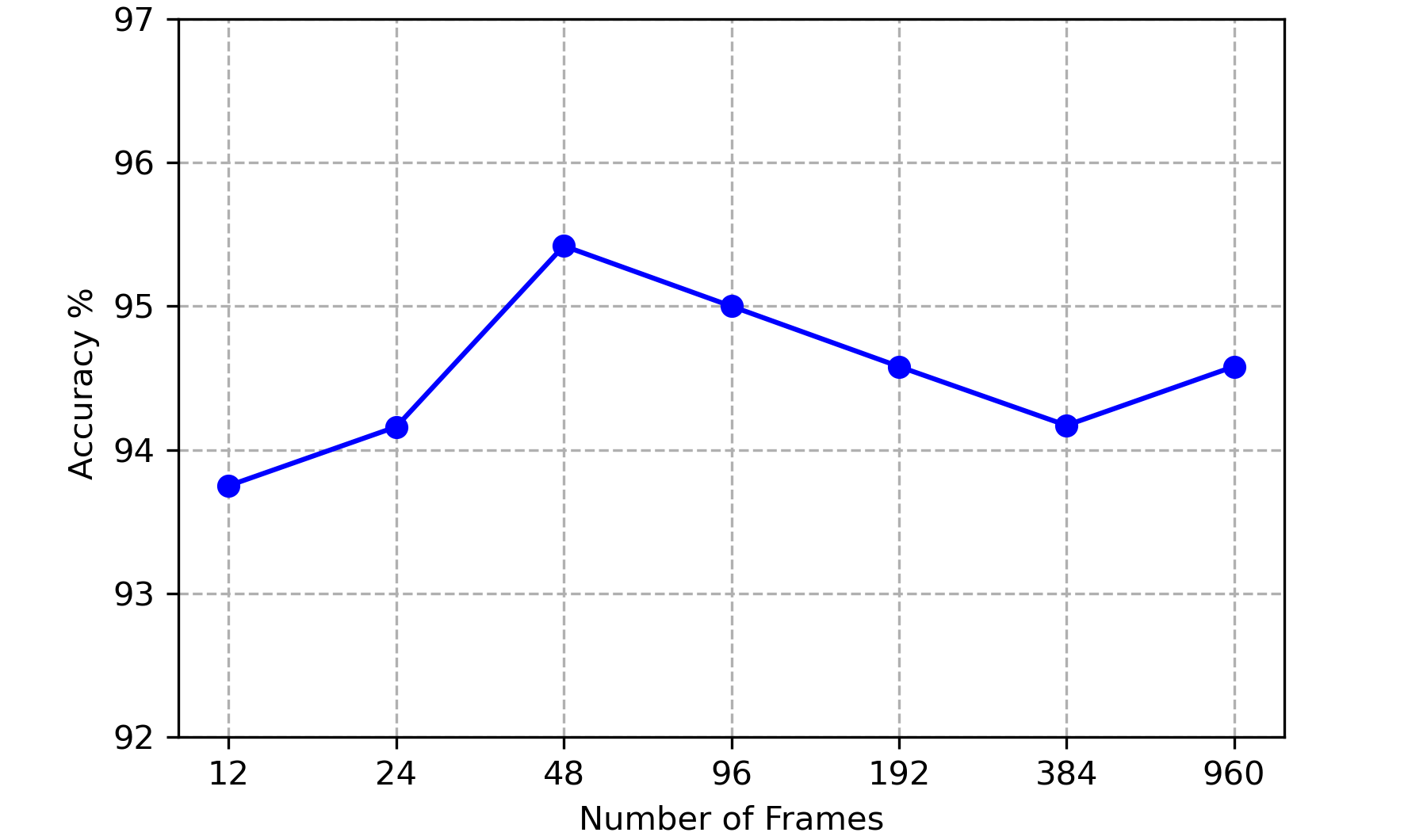}
\caption{Evolution of model accuracy based on the number of frames.}
\label{fig:accuracyPerNbOfFrames}
\end{center}
\end{figure}

\textbf{frames Number:}
In Fig.\ref{fig:accuracyPerNbOfFrames}, we reveal the impact of different numbers of frames on the performance of the transformer model. We evaluate the model using various numbers of frames and recorded the corresponding accuracies.
Initially, as the number of frames increased from '12' to '48', the accuracy steadily improved, reaching its peak at '48' frames with an accuracy of $95.42\%$. Beyond this point, with further increases in the number of frames, the accuracy exhibit a slight fluctuation but remained consistently high, indicating the model robustness in handling longer sequences.\\

\textbf{attention head Number:}
The performance of the transformer model is largely influenced by the number of attention heads used. To investigate this connection, We examine the effect of varying the number of multi-head self-attention heads on the temporal transformer architecture. Fig.\ref{fig:accuracyPerNbOfFrames} reports the evolution of the model accuracy depending on the number of heads. The figure demonstrates an initial enhancement in accuracy as the number of heads is augmented from [1, 1, 1] to [2, 2, 2]. However,  beyond this point, the accuracy reaches a plateau and even diminishes with further increases in attention heads. This behavior could be explained by the positive correlation between the number of attention heads and the model capacity to capture complex patterns and dependencies. Consequently, in scenarios where the complexity and scale of the data are insufficient, an elevated susceptibility to overfitting becomes a potential outcome.

\begin{figure}[h]
\begin{center}
\includegraphics[scale=0.15]{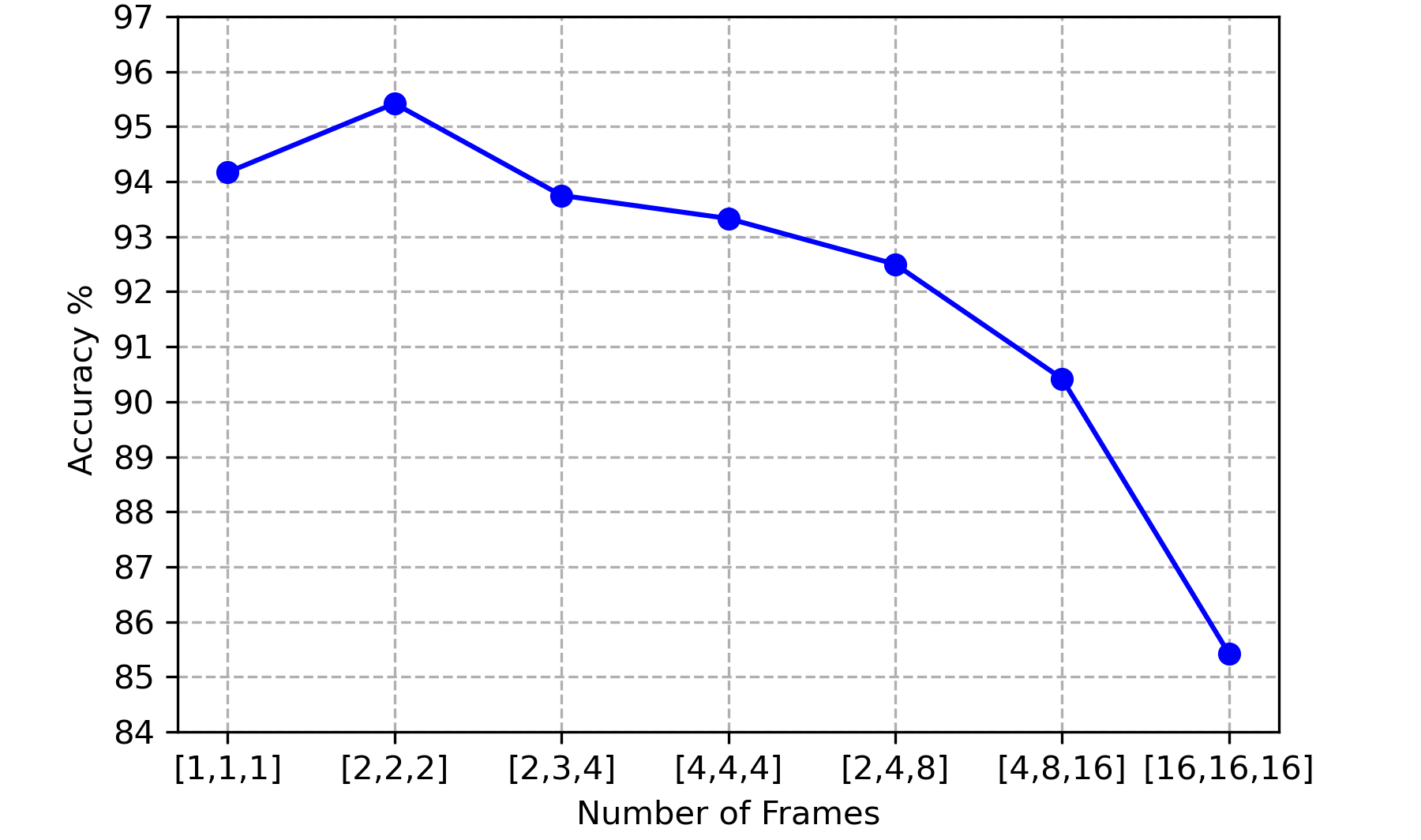}
\caption{Evolution of model accuracy based on the number of multi-head self-attention heads.}
\label{fig:accuracyPerNbOfHeads}
\end{center}
\end{figure}

\subsection{Discussion}
Our experiments prove that the proposed model shows competitive results to state-of-the-art. This is due to the use of a combination of \textbf{1)} spiral convolutions to exploit the geometric information directly off the mesh object, and \textbf{2)} self-attention, employed in the transformer, to model the temporal evolution of the sequence. 
In addition, the proposed model proved to be very light compared to the point cloud-based methods. when trained on sequences of the same length, our model shows more memory efficiency, using only $2\%$ - $15\%$ of the memory used by the baselines. This, combined with the use of transformers that can handle large sequences, enables our model to process videos of more $1000+$ frames without any memory issues. Moreover, the proposed model offers more resource efficiency by extracting task-free features that can be used for multiple purposes without the need to retrain the feature extractor. 

However, there are some challenges that flaw our SpATr model. The main limitation is that the proposed model requires meshes of the same topology which calls for MoCap data, in contrast to point cloud-based methods that can handle the irregularity of 3D scans directly. Additionally, decoupling the action recognition, first, results in the extraction of task-free features that contain shape, action, pose, and appearance, and stacks the errors of both sub-networks. This has the potential to negatively impact the accuracy of the model. Moreover, due to their capability to effectively handle long-range dependencies, transformers hold great promise in HAR, however, they require large amounts of high-quality data. 

\section{Conclusion And Perspectives}
\label{conclusion}
In this paper, we present a novel 3D HAR model that decouples the task into spatial and temporal modeling. Specifically, we address the problem of high resource consumption encountered by state-of-the-art methods, by utilizing lightweight spiral convolutions to encode the 3D mesh and a tiny temporal transformer to capture temporal evolution. We have deeply evaluated our method on Babel, MoVi, and BMLrub databases, by presenting the accuracy and the memory usage of our model and the state-of-the-art in different settings. The results of this evaluation confirm our claim that our method is competitive with the baselines in recognition accuracy significantly outperforming them in memory efficiency. 


\bibliographystyle{model2-names}
\bibliography{refs.bib}
\clearpage
\onecolumn

\appendix
\section{Confusion Matrices}
\begin{figure*}[h!]
    \centering
    
    \begin{subfigure}{0.48\linewidth}
        \centering
        \includegraphics[scale=0.37]{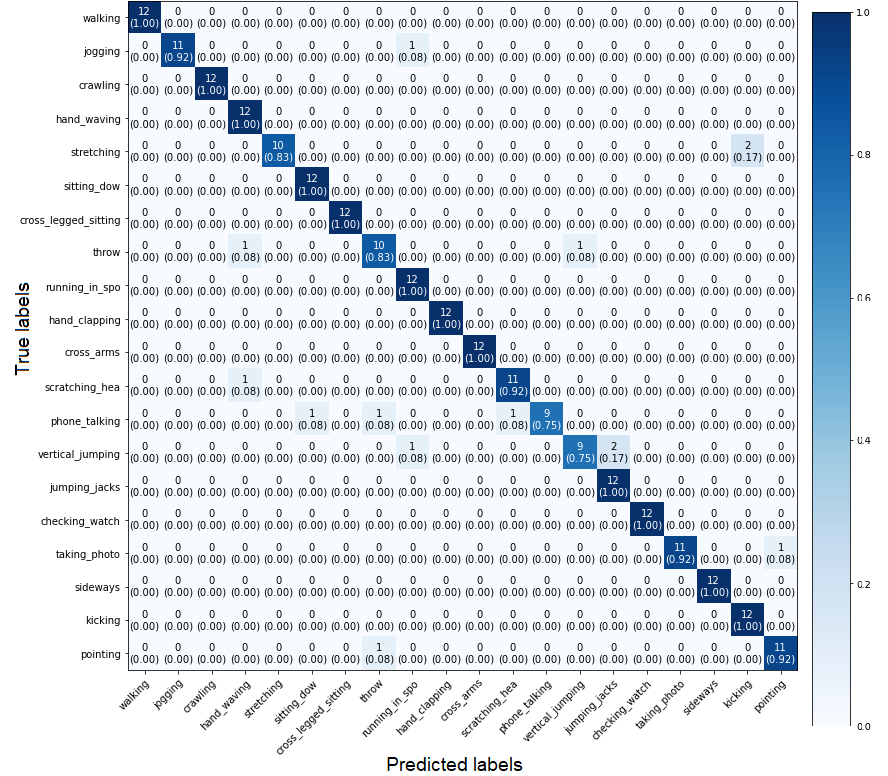}
        \caption{MoVi}
        \label{subfig:movi_res}
    \end{subfigure}
    \hfill
    \begin{subfigure}{0.5\linewidth}
        \centering
        \includegraphics[scale=0.37]{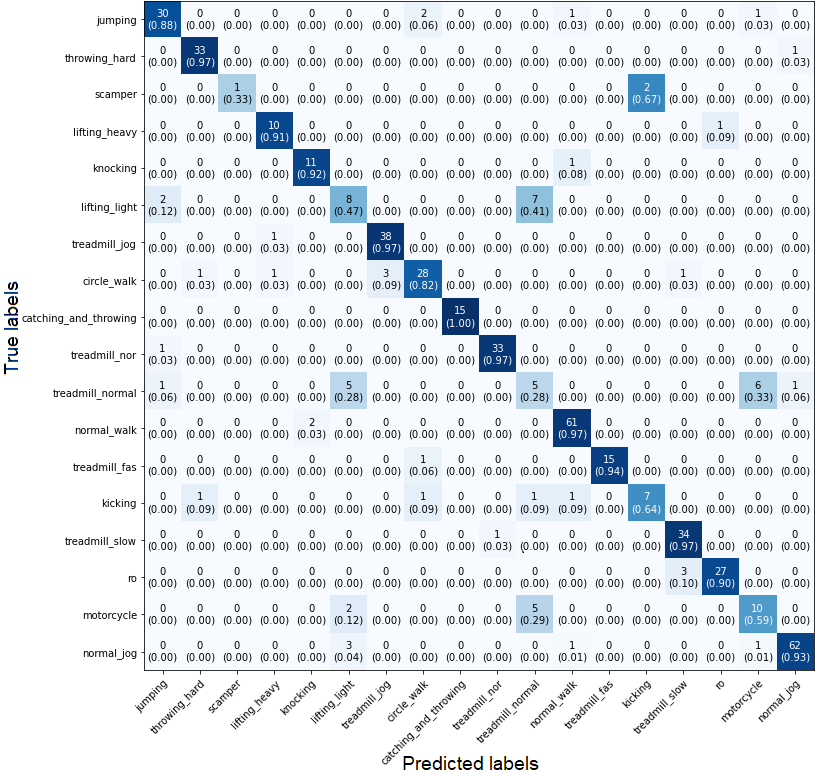}
        \caption{BMLrub}
        \label{subfig:bml_res}
    
    \end{subfigure}
\begin{subfigure}{0.7\linewidth}
        \centering
        \includegraphics[width=\linewidth]{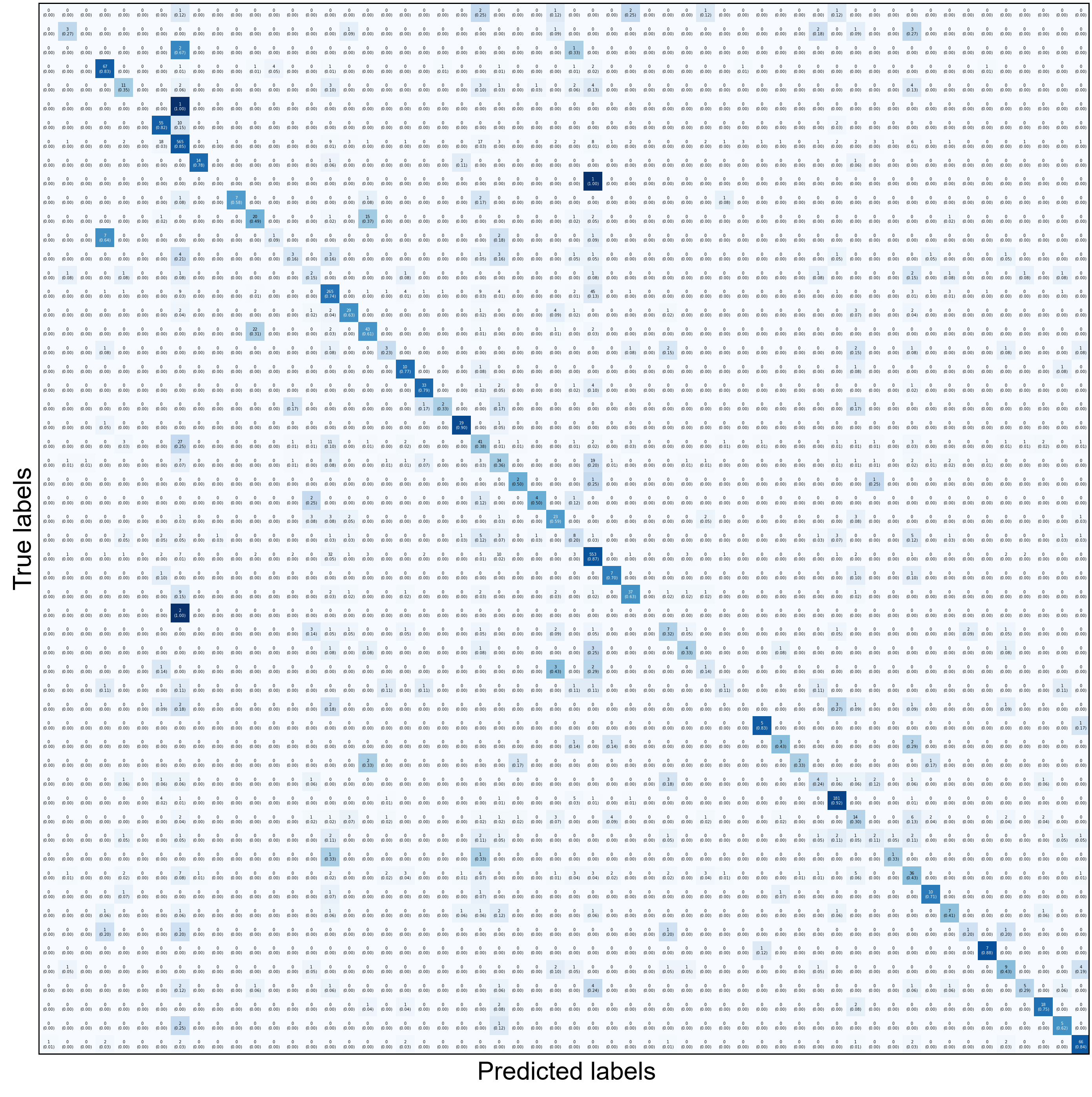}
        \caption{Babel}
        \label{subfig:babel_res}
    
    \end{subfigure}
    \caption{confusion matrices showing the classification results of our proposed model on the test set of (a) MoVi, (b) BMLrub, and (c) Babel databases. We note that the model
    is trained on 24 frames.}
    \label{fig:confusion_matrices}
\end{figure*}

\end{document}